\documentclass[runningheads]{llncs}
\usepackage{graphicx}
\usepackage{multirow}

\usepackage{color}
\newcommand{\dg}[1]{\textcolor{black}{#1}}

\begin{document}
%
\title{PseudoEdgeNet: Nuclei Segmentation \\ only with Point Annotations}
%
%
\author{Inwan Yoo \and Donggeun Yoo \and Kyunghyun Paeng}
\authorrunning{Yoo et al.}
%
\institute{Lunit Inc., Seoul, South Korea \\
\email{\{iwyoo,dgyoo,khpaeng\}@lunit.io}}
\maketitle              
\begin{abstract}
Nuclei segmentation is one of the important tasks for whole slide image analysis in digital pathology. With the drastic advance of deep learning, recent deep networks have demonstrated successful performance of the nuclei segmentation task. However, a major bottleneck to achieving good performance is the cost for annotation. A large network requires a large number of segmentation masks, and this annotation task is given to pathologists, not the public. In this paper, we propose a weakly supervised nuclei segmentation method, which requires only point annotations for training. This method can scale to large training set as marking a point of a nucleus is much cheaper than the fine segmentation mask. To this end, we introduce a novel auxiliary network, called PseudoEdgeNet, which guides the segmentation network to recognize nuclei edges even without edge annotations. We evaluate our method with two public datasets, and the results demonstrate that the method consistently outperforms other weakly supervised methods.

\keywords{Nuclei segmentation \and Weakly supervised learning \and Point annotation.}
\end{abstract}
\section{Introduction}
%
With the advent of digital pathology~\cite{al2012digital}, extracting information of biological components from whole slide images (WSIs) is attracting more attention since the statistics can be utilized for biomarker development as well as accurate diagnosis~\cite{beck2011systematic}. However, it is infeasible for human experts (e.g. pathologists) to manually extract the statistics due to the huge dimensions of WSI space. A WSI can comprise up to 100k$\times$100k pixels~\cite{liu2017detecting}. Despite its huge dimensions, the area of a target instance is usually small, such as a tumor cell. In order to automate this process, a variety of visual recognition methods from computer vision has been applied to WSIs~\cite{litjens2016deep,zhou2018sfcn,kumar2017dataset,naylor2018segmentation}. Among the various recognition tasks, this paper focuses on the nuclei segmentation problem~\cite{naylor2018segmentation}.

During the last few years, we have witnessed drastic progress in segmentation tasks on WSIs \cite{kumar2017dataset,naylor2018segmentation} with deep learning. Despite its successful performance, the cost for annotations is still worrisome. Drawing fine masks of target instances is much more labor-intensive than drawing bounding boxes or tagging class labels. Furthermore, only experts such as pathologists, not the public, can conduct this annotation task. The situation gets much worse when we choose a deep network as a segmentation model which can be learned with a large number of training samples. These factors make it difficult to create a large-scale segmentation dataset in the WSI domain.
%

This paper aims at cutting the annotation cost for nuclei segmentation. The quickest and easiest way to annotate a nucleus is to mark a point on it. A point does not contain fine boundary information of a nucleus, but we can obtain a much larger amount of training samples than segmentation masks, given a fixed budget for annotation. This strategy is scalable for learning a large network, and it is also expected that a large amount of training samples will contribute to the generalization performance~\cite{mahajan2018exploring} of the network.

To this end, we propose a novel weakly-supervised model, which is composed of a segmentation network and an auxiliary network, called PseudoEdgeNet. The segmentation network produces nuclei segments while the auxiliary network helps the main network learn to recognize nuclei boundaries with point annotations only. We evaluate this model over two public datasets~\cite{kumar2017dataset,naylor2018segmentation} and the results demonstrate successful segmentation performance compared to other recent methods~\cite{laradji2018blobs,tang2018regularized} for weakly-supervised segmentation.

\section{Related Research}
\subsubsection{Nuclei segmentation}
There have been several works for nuclei segmentation based on deep learning, but all of the methods use a fully-supervised learning model that requires nuclei segmentation masks.
\cite{kumar2017dataset} makes a public nuclei dataset containing full segmentation masks and introduces a segmentation model based on a pixel-level classification approach.
\cite{naylor2018segmentation} approaches the nuclei segmentation task as a regression problem.
The work done by~\cite{zhou2015nuclei} is also a regression method but a sparsity constraint is introduced.
\cite{akram2016cell} adopts a two-step approach where the model produces cell proposals first and then segments the nuclei.

\subsubsection{Cell detection with points}
Cell detection methods are related to ours since these often utilize point annotations~\cite{zhou2018sfcn,kainz2015you,weidi2015microscopy}.
One popular family casts cell detection as a regression problem, such as~\cite{kainz2015you} and~\cite{weidi2015microscopy} adopt a regression Random Forest and a CNN regressor, respectively.
Another approach is pixel-level classification with point annotations~\cite{zhou2018sfcn}, which is similar to the typical semantic segmentation approach.
However, these methods use the point annotations to learn a detection model which predicts the cell locations as points, not a segmentation model.


\subsubsection{Weakly-supervised segmentation}
To the best of our knowledge, there has been no weakly-supervised method for nuclei segmentation. However, in the natural image domain, a long line of works has been presented to reduce the cost of pixel-level annotations.
An object segmentation model is learned with bounding-boxes~\cite{khoreva2017simple} or scribbles~\cite{lin2016scribblesup,tang2018regularized}, which are much cheaper to obtain than the pixel-level masks.
The work presented by~\cite{laradji2018blobs} is similar to ours in that it also uses point annotations. However, their target task is to find ``rough blobs'' on objects while we have to predict ``fine boundaries'' of nuclei.


\section{Method}
\begin{figure}[t]
\centering
\includegraphics[width=0.9\textwidth]{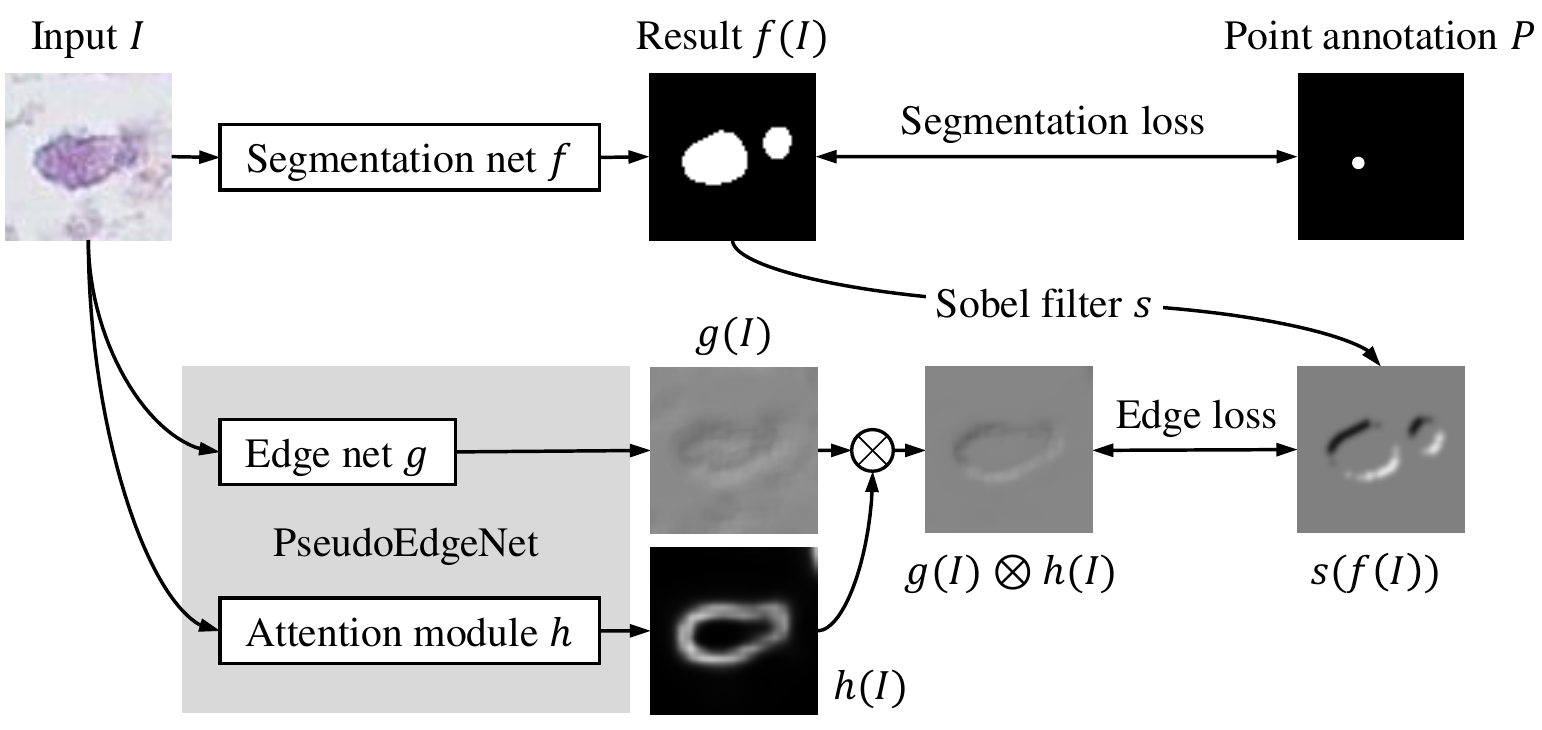}
\caption{The overall architecture for weakly-supervised nuclei segmentation. The segmentation network $f$ is jointly learned with PseudoEdgeNet $\{g, h\}$. In edge maps, the gray color represents zero while the white and black colors encode positive and negative pixel values, respectively.}
\label{fig:architecture}
\end{figure}

The proposed architecture is composed of a segmentation network and PseudoEdgeNet. The segmentation network is our target model that segments nuclei from inputs. PseudoEdgeNet, only introduced for training phase, encourages the segmentation network to recognize nucleus edges without edge annotations. Figure~\ref{fig:architecture} is illustrating the proposed architecture.

\subsection{Segmentation Network}
\label{sec:method:segmentation_network}

To learn a segmentation network with point annotations, we follow the label assignment scheme presented by~\cite{laradji2018blobs}. In this scheme, positive labels are given to the pixels corresponding to point annotations, while negative labels are assigned to pixels on Voronoi boundaries that can be obtained by distance transform with point annotations. Then, binary cross-entropy losses are evaluated and averaged over the labels and corresponding pixel outputs.

The segmentation network learned with this loss can successfully localize nuclei as blobs. However, it fails to segment along the edges of nuclei since there is no direct supervision for edges. For this reason, we introduce an auxiliary network that can provide fine boundary information with that the segmentation network is supervised to segment along the nucleus edges.

\subsection{Learning with PseudoEdgeNet}
\label{section:method:learning_with_an_edge_network}

In CNNs, it is well known that lower layers extract low-level information such as edges and blobs, while higher layers encode object parts or an object as a whole~\cite{zeiler2014visualizing}. This motivates us to design a shallow CNN to efficiently extract nucleus edges \textit{without} edge annotations. These pseudo edges can be inaccurate but sufficient to act as supervisory signals to the segmentation network.

Given an image $I$, since we do not have edge annotations, PseudoEdgeNet $g$ is jointly learned with the segmentation network $f$ using the point annotations $P$. To make the edge map $g(I)$ comparable to the segmentation map $f(I)$, we apply a ($x,y$)-directional Sobel filter $s$ to $f(I)$. Then, the final loss $\mathcal{L}$ to jointly learn these two networks $\{f,g\}$ is defined as
\begin{equation}
\mathcal{L}(I, P, f, g)=\mathcal{L}_\textrm{ce}(f(I), P) + \lambda\cdot|\textrm{s}(f(I))-g(I)|,
\label{eq:loss}
\end{equation}
where $\mathcal{L}_\textrm{ce}$ is the pixel-averaged cross-entropy loss defined in Section~\ref{sec:method:segmentation_network} and $\lambda$ is a scaling constant. The segmentation network $f$ is learned to detect nuclei by the first term, and simultaneously forced to activate on nucleus edges by the second term. PseudoEdgeNet $g$ is used only to learn $f$ with this loss, and unnecessary at inference time.

What is noteworthy here is the capacity gap between $f$ and $g$. If $g$ is as large as $f$, then $g$ will be learned just like $f$, except that the outputs are edges. However, since we design $g$ to be much smaller than $f$, $g$ is able to encode low/mid-level edges, not the high-level information, which only $f$ can cope with. Empirical analysis on this will be discussed later with Table~\ref{tab:ncg} in Section~\ref{sec:evaluation:results}.

\subsection{Attention Module for Edge Network}
\begin{figure}[t]
\centering
\begin{tabular}{ccccccc}
\includegraphics[height=1.4cm]{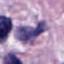}&
\includegraphics[height=1.4cm]{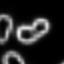}&
\includegraphics[height=1.4cm]{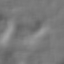}&
\includegraphics[height=1.4cm]{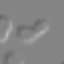}&
\includegraphics[height=1.4cm]{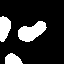}&
\includegraphics[height=1.4cm]{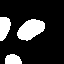}&
\includegraphics[height=1.4cm]{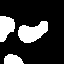}\\
\includegraphics[height=1.4cm]{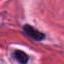}&
\includegraphics[height=1.4cm]{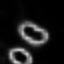}&
\includegraphics[height=1.4cm]{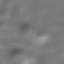}&
\includegraphics[height=1.4cm]{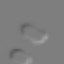}&
\includegraphics[height=1.4cm]{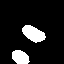}&
\includegraphics[height=1.4cm]{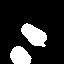}&
\includegraphics[height=1.4cm]{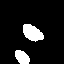}\\
\includegraphics[height=1.4cm]{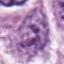}&
\includegraphics[height=1.4cm]{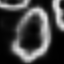}&
\includegraphics[height=1.4cm]{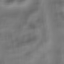}&
\includegraphics[height=1.4cm]{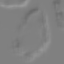}&
\includegraphics[height=1.4cm]{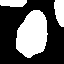}&
\includegraphics[height=1.4cm]{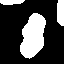}&
\includegraphics[height=1.4cm]{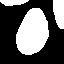}\\
\includegraphics[height=1.4cm]{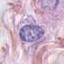}&
\includegraphics[height=1.4cm]{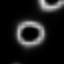}&
\includegraphics[height=1.4cm]{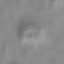}&
\includegraphics[height=1.4cm]{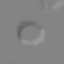}&
\includegraphics[height=1.4cm]{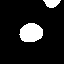}&
\includegraphics[height=1.4cm]{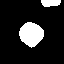}&
\includegraphics[height=1.4cm]{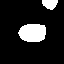}\\
\includegraphics[height=1.4cm]{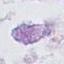}&
\includegraphics[height=1.4cm]{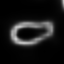}&
\includegraphics[height=1.4cm]{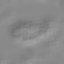}&
\includegraphics[height=1.4cm]{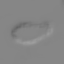}&
\includegraphics[height=1.4cm]{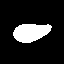}&
\includegraphics[height=1.4cm]{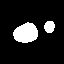}&
\includegraphics[height=1.4cm]{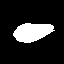}\\
\includegraphics[height=1.4cm]{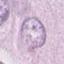}&
\includegraphics[height=1.4cm]{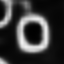}&
\includegraphics[height=1.4cm]{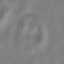}&
\includegraphics[height=1.4cm]{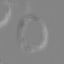}&
\includegraphics[height=1.4cm]{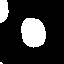}&
\includegraphics[height=1.4cm]{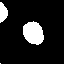}&
\includegraphics[height=1.4cm]{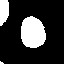}\\
(a) $I$&(b) $h(I)$&(c) $g(I)$&(d) $g \otimes h$&(e) Ours&(f)~\cite{laradji2018blobs}&(g) GT
\end{tabular}
\caption{Qualitative examples and comparisons: (a) inputs, (b) attention maps, (c) ($x,y$)-directional raw edge maps, (d) final edge maps in which attentions are multiplied, (e) final segmentation results from our segmentation network, (f) segmentation results from the baseline method~\cite{laradji2018blobs}, and (g) ground-truth masks. In (c, d), each map is averaging the $x$- and $y$-directional edge maps. The gray color represents zero while the white and black colors encode positive and negative pixel values, respectively.}
\label{fig:qualitative_comparison}
\end{figure}

According to our experiment in Table~\ref{tab:performance}, the method presented up to Section~\ref{section:method:learning_with_an_edge_network} shows clear performance gains. However, there is still much room for improving the quality of edges used for auxiliary supervision. Due to the low capacity of $g$, a significant portion of edges originates from irrelevant backgrounds. To suppress these, we add an attention module $h$ inside PseudoEdgeNet, which produces an attention map $h(I)$, that indicates where to extract edges. Since this task requires high-level understanding of nuclei, we use a large architecture for this module. The attention map $h(I)$ is applied to the raw edge $g(I)$, and then the loss function is re-defined as
\begin{equation}
\mathcal{L}(I, P, f, g, h)=\mathcal{L}_\textrm{ce}(f(I), P) + \lambda\cdot|\textrm{s}(f(I))-g(I)\otimes h(I)|,
\label{eq:loss_with_attention}
\end{equation}
where $\otimes$ means element-wise multiplication. We jointly learn parameters in $\{f, g, h\}$, and only use the segmentation network $f$ at inference time. Figure~\ref{fig:qualitative_comparison}\dg{-(b, c, d)} shows how attention improves quality of edges.

\section{Evaluation}

\subsection{Datasets}
We evaluate our method with two major nuclei segmentation datasets: MoNuSeg~\cite{kumar2017dataset} and {\itshape TNBC}~\cite{naylor2018segmentation}.
MoNuSeg comprises 30 images in which each image size is 1,000$\times$1,000. TNBC is composed of 50 images with 512$\times$512 size. These two datasets provide full nuclei masks, that enable us to automatically generate point annotations and to evaluate segmentation results with full masks.
To construct a training set composed of images and point annotations, we extract nuclei points by calculating the center of mass of each nucleus instance mask.
We conduct $k$-fold cross-validation with $k$=$10$ for thorough evaluation. Among 10 folds of data, we use two folds as a validation set and a test set, and the rest as a training set.


\subsection{Implementation Details}

We choose \cite{laradji2018blobs} as a baseline method, which is the most recent work for learning with point annotations. Among the loss terms in~\cite{laradji2018blobs}, we do not use the image-level classification loss since almost of image patches contain nuclei, but only adopt the pixel-averaged cross-entropy loss $\mathcal{L}_{ce}$ described in Section~\ref{sec:method:segmentation_network}.

We employ a Feature Pyramid Network (FPN) for segmentation~\cite{kirillov2017unified} with a ResNet-50~\cite{he2016deep} backbone followed by a sigmoid layer as the segmentation network $f$ in all experiments.
We compose $g$ of PseudoEdgeNet with four convolution layers to make it much smaller than the segmentation network $f$. Each of the convolution layers contains 64 filters and is followed by batch normalization and ReLU, except for the last layer, which produces a two-channel output representing ($x,y$)-directional Sobel edge maps. For the attention module $h$ inside PseudoEdgeNet, we use an FPN with a Resnet-18 backbone and stack a sigmoid as an output layer.

We set the label weights applied to the cross-entropy loss as 0.1 and 1.0 for negative and positive labels respectively since much more negatives are given to the loss function compared to the positives. The scaling constant $\lambda$ in Equation~(\ref{eq:loss}) and (\ref{eq:loss_with_attention}) is set to 1.0.

To achieve high generalization performance, we apply a lot of data augmentation methods to inputs including color jittering, Gaussian blurring, Gaussian noise injection, rotation, horizontal or vertical flip, affine transformation, and elastic deformation.
We use an Adam optimizer with an initial learning rate of 0.001. We train all networks with a plateau scheduling policy where the learning rate is halved when the average loss per epoch does not decrease for the current five epochs.

The threshold to determine positive pixels from $f(I)$ is 0.5. We evaluate the model on the validation set for each epoch and choose the best model to evaluate that on the test set. We choose the intersection over union (IoU), which is the most common metric for semantic segmentation, as an evaluation metric.

\subsection{Results}
\label{sec:evaluation:results}


\begin{table}[t]
\caption{Nuclei segmentation performance comparison between methods. The mean and standard deviation of 10-fold cross-validation results (10 IoU scores) are reported.}
\centering
\begin{tabular}{|l|c|c|}
\hline
Methods              & MoNuSeg       & TNBC          \\ \hline\hline
Baseline \cite{laradji2018blobs} & 0.5710 ($\pm$0.02) & 0.5504 ($\pm$0.04) \\
DenseCRF* \cite{tang2018regularized} & 0.5813 ($\pm$0.03) & 0.5555 ($\pm$0.04) \\
PseudoEdgeNet with large $g$ & 0.5786 ($\pm$0.04) & 0.5787 ($\pm$0.04) \\
PseudoEdgeNet with small $g$& 0.6059 ($\pm$0.04) & 0.5853 ($\pm$0.03) \\
PseudoEdgeNet with small $g$ and $h\;$& \textbf{0.6136} ($\pm$0.04) & \textbf{0.6038} ($\pm$0.03) \\ \hline\hline
Fully supervised (upper bound)& 0.6522 ($\pm$0.03) & 0.6619 ($\pm$0.04) \\ \hline
\multicolumn{3}{l}{*Authors' open source is used: https://github.com/meng-tang/rloss}
\end{tabular}
\label{tab:performance}
\end{table}

Table~\ref{tab:performance} is summarizing the experimental results. The most recent weakly supervised segmentation method~\cite{tang2018regularized} noted by DenseCRF marginally beats the baseline~\cite{laradji2018blobs}. However, PseudoEdgeNet with small $g$ significantly improves the baseline by a large margin of +3.49\% for both of the datasets. When PseudoEdgeNet is equipped with the attention module $h$, the performance gains increase to +4.26\% and +5.34\%. These results clearly demonstrate the effectiveness of $g$ and $h$. Compared to small $g$, the worse performance of large $g$ proves the importance of the capacity of $g$. For large $g$, we use the same architecture of $f$, so $g$ is learned just like $f$, resulting in a small improvement to the baseline.

To take a closer look at the importance of capacity, we try to densely change the depth of $g$. Table~\ref{tab:ncg} is summarizing the results. For the family of small networks, the depth change from 2 to 8 does not make any significant difference in performance. However, when $g$ is equipped with large ResNets, the performance significantly drops. The depth variation within the large architecture also shows minor performance changes.

\begin{table}[t]
\caption{Nuclei segmentation performance to the size of edge networks. The mean and standard deviation of 10-fold cross-validation results (10 IoU scores) are reported.} \label{tab:ncg}
\centering
\begin{tabular}{|c|l|c|c|}
\hline
\multicolumn{2}{|l|}{Edge networks ($g$)}& MoNuSeg       & TNBC          \\ \hline\hline
\multirow{4}{*}{Small}&CNN with 2 conv layers   & 0.6117 ($\pm$0.03) & 0.5928 ($\pm$0.04) \\
&CNN with 4 conv layers   & \textbf{0.6136} ($\pm$0.04) & \textbf{0.6038} ($\pm$0.03) \\
&CNN with 6 conv layers   & 0.6105 ($\pm$0.04) & 0.5896 ($\pm$0.03) \\
&CNN with 8 conv layers   & 0.6119 ($\pm$0.02) & 0.5934 ($\pm$0.04) \\ \hline\hline
\multirow{3}{*}{Large}&FPN-ResNet18  & 0.6005 ($\pm$0.03) & 0.5795 ($\pm$0.04) \\
&FPN-ResNet34  & 0.6069 ($\pm$0.03) & 0.5796 ($\pm$0.03) \\
&FPN-ResNet50  & 0.5786 ($\pm$0.04) & 0.5787 ($\pm$0.04) \\ \hline
\end{tabular}
\end{table}

\section{Conclusion and Future Work}
 We have presented a novel nuclei segmentation method only with point supervision. Our auxiliary network, PseudoEdgeNet, can find object edges without edge annotations as it has low capacity, which acts as a strong constraint for weakly-supervised learning. Our method can scale to large-scale segmentation problem for better performance, as point annotations are much cheaper than segmentation masks.
 
 However, given the same amount of data, the performance of weakly-supervised learning is bounded to that of supervised learning. It will be a promising future work to annotate a small number of segmentation masks, and use both mask and point annotations to achieve performance comparable to supervised learning, while greatly saving the annotation cost.

\bibliographystyle{splncs04}
\bibliography{reference}

\begin{thebibliography}{10}
\providecommand{\url}[1]{\texttt{#1}}
\providecommand{\urlprefix}{URL }
\providecommand{\doi}[1]{https://doi.org/#1}

\bibitem{akram2016cell}
Akram, S.U., Kannala, J., Eklund, L., Heikkil{\"a}, J.: Cell segmentation
  proposal network for microscopy image analysis. In: Deep Learning and Data
  Labeling for Medical Applications, pp. 21--29. Springer (2016)

\bibitem{al2012digital}
Al-Janabi, S., Huisman, A., Van~Diest, P.J.: Digital pathology: current status
  and future perspectives. Histopathology  \textbf{61}(1), ~1--9 (2012)

\bibitem{beck2011systematic}
Beck, A.H., Sangoi, A.R., Leung, S., Marinelli, R.J., Nielsen, T.O., Van
  De~Vijver, M.J., West, R.B., Van De~Rijn, M., Koller, D.: Systematic analysis
  of breast cancer morphology uncovers stromal features associated with
  survival. Science translational medicine  \textbf{3}(108),
  108ra113--108ra113 (2011)

\bibitem{he2016deep}
He, K., Zhang, X., Ren, S., Sun, J.: Deep residual learning for image
  recognition. In: Proceedings of the IEEE conference on computer vision and
  pattern recognition (2016)

\bibitem{kainz2015you}
Kainz, P., Urschler, M., Schulter, S., Wohlhart, P., Lepetit, V.: You should
  use regression to detect cells. In: International Conference on Medical Image
  Computing and Computer-Assisted Intervention. pp. 276--283. Springer (2015)

\bibitem{khoreva2017simple}
Khoreva, A., Benenson, R., Hosang, J., Hein, M., Schiele, B.: Simple does it:
  Weakly supervised instance and semantic segmentation. In: Proceedings of the
  IEEE conference on computer vision and pattern recognition. pp. 876--885
  (2017)

\bibitem{kirillov2017unified}
Kirillov, A., He, K., Girshick, R., Doll{\'a}r, P.: A unified architecture for
  instance and semantic segmentation

\bibitem{kumar2017dataset}
Kumar, N., Verma, R., Sharma, S., Bhargava, S., Vahadane, A., Sethi, A.: A
  dataset and a technique for generalized nuclear segmentation for
  computational pathology. IEEE transactions on medical imaging
  \textbf{36}(7),  1550--1560 (2017)

\bibitem{laradji2018blobs}
Laradji, I.H., Rostamzadeh, N., Pinheiro, P.O., Vazquez, D., Schmidt, M.: Where
  are the blobs: Counting by localization with point supervision. In:
  Proceedings of the European Conference on Computer Vision (ECCV). pp.
  547--562 (2018)

\bibitem{lin2016scribblesup}
Lin, D., Dai, J., Jia, J., He, K., Sun, J.: Scribblesup: Scribble-supervised
  convolutional networks for semantic segmentation. In: Proceedings of the IEEE
  Conference on Computer Vision and Pattern Recognition. pp. 3159--3167 (2016)

\bibitem{litjens2016deep}
Litjens, G., S{\'a}nchez, C.I., Timofeeva, N., Hermsen, M., Nagtegaal, I.,
  Kovacs, I., Hulsbergen-Van De~Kaa, C., Bult, P., Van~Ginneken, B., Van
  Der~Laak, J.: Deep learning as a tool for increased accuracy and efficiency
  of histopathological diagnosis. Scientific reports  \textbf{6},  26286 (2016)

\bibitem{liu2017detecting}
Liu, Y., Gadepalli, K., Norouzi, M., Dahl, G.E., Kohlberger, T., Boyko, A.,
  Venugopalan, S., Timofeev, A., Nelson, P.Q., Corrado, G.S., et~al.: Detecting
  cancer metastases on gigapixel pathology images. arXiv preprint
  arXiv:1703.02442  (2017)

\bibitem{mahajan2018exploring}
Mahajan, D., Girshick, R., Ramanathan, V., He, K., Paluri, M., Li, Y.,
  Bharambe, A., van~der Maaten, L.: Exploring the limits of weakly supervised
  pretraining. In: The European Conference on Computer Vision (ECCV) (September
  2018)

\bibitem{naylor2018segmentation}
Naylor, P., La{\'e}, M., Reyal, F., Walter, T.: Segmentation of nuclei in
  histopathology images by deep regression of the distance map. IEEE
  Transactions on Medical Imaging  (2018)

\bibitem{tang2018regularized}
Tang, M., Perazzi, F., Djelouah, A., Ben~Ayed, I., Schroers, C., Boykov, Y.: On
  regularized losses for weakly-supervised cnn segmentation. In: Proceedings of
  the European Conference on Computer Vision (ECCV). pp. 507--522 (2018)

\bibitem{weidi2015microscopy}
Weidi, X., Noble, J.A., Zisserman, A.: Microscopy cell counting with fully
  convolutional regression networks. In: 1st Deep Learning Workshop, Medical
  Image Computing and Computer-Assisted Intervention (MICCAI) (2015)

\bibitem{zeiler2014visualizing}
Zeiler, M.D., Fergus, R.: Visualizing and understanding convolutional networks.
  In: European conference on computer vision. pp. 818--833. Springer (2014)

\bibitem{zhou2018sfcn}
Zhou, Y., Dou, Q., Chen, H., Qin, J., Heng, P.A.: Sfcn-opi: Detection and
  fine-grained classification of nuclei using sibling fcn with objectness prior
  interaction. In: Thirty-Second AAAI Conference on Artificial Intelligence
  (2018)

\bibitem{zhou2015nuclei}
Zhou, Y., Chang, H., Barner, K.E., Parvin, B.: Nuclei segmentation via sparsity
  constrained convolutional regression. In: 2015 IEEE 12th International
  Symposium on Biomedical Imaging (ISBI). pp. 1284--1287. IEEE (2015)

\end{thebibliography}

\end{document}